\documentclass{egpubl}
\usepackage{amsmath}
\usepackage{makecell}
\usepackage{multirow}
\usepackage{multicol}
\usepackage{float}

\JournalPaper         
\CGFccby

\usepackage[T1]{fontenc}
\usepackage{dfadobe}  

\usepackage{cite}  
\usepackage[font=normalsize,labelfont=bf, aboveskip=4pt, belowskip=4pt]{subcaption}

\usepackage{enumitem}
\usepackage{calc}
\BibtexOrBiblatex
\electronicVersion
\PrintedOrElectronic

\usepackage{graphicx}
\usepackage{sidecap}
\usepackage{egweblnk}
\usepackage{tikz}
\usetikzlibrary{patterns,shapes.arrows,calc}
\usetikzlibrary{positioning}
\usetikzlibrary{arrows.meta}
\usepackage{pgfplots}
\pgfplotsset{compat=1.18}
\usepackage{soul}

\usepackage{ifthen}
\newboolean{revising}
\setboolean{revising}{false}
\ifthenelse{\boolean{revising}}
{
    
} {
    
}

\newif\ifwarntodo 
\warntodotrue

\newif\ifshowtodo
\showtodotrue

\ifwarntodo
	
\else
	
\fi

\makeatletter
\renewcommand{\section}{\@startsection{section}{1}{\z@}%
  {1.0ex}
  {0.01ex \@plus 0ex \@minus 0ex}
  {\normalfont\normalsize\bfseries}}
\renewcommand{\subsection}{\@startsection{subsection}{1}{\z@}%
  {-0.5ex}
  {0ex}
  {\normalfont\normalsize\bfseries}}
\makeatother

\newcommand{\changed}[1]{#1}
\newcommand{\allchanged}{}
\newcommand{\alladded}{}
\newcommand{\allsame}{}

\renewcommand{\st}[1]{}

\title[Interactive Groupwise Comparison for RLHF]%
      {Interactive Groupwise Comparison for \\Reinforcement Learning from Human Feedback}

\author[J. Kompatscher, D. Shi, G. Varni, T. Weinkauf \& A. Oulasvirta]
 {\parbox{\textwidth}{\centering 
 Jan Kompatscher$^{1}$, 
 Danqing Shi$^{1,2}$, 
 Giovanna Varni$^{3}$, 
 Tino Weinkauf$^{4}$, 
 and Antti Oulasvirta$^{1}$}
 \\
 {\parbox{\textwidth}{\centering 
 $^1$Aalto University, Finland\\
 $^2$University of Cambridge, United Kingdom\\
 $^3$University of Trento, Italy\\
 $^4$KTH Royal Institute of Technology, Sweden
 }
 }
 }

\begin{document}

\newlength{\leftteaserwidth}
\setlength{\leftteaserwidth}{\textwidth}
\newlength{\rightteaserwidth}
\setlength{\rightteaserwidth}{\textwidth}
\addtolength{\rightteaserwidth}{-\leftteaserwidth}
\addtolength{\rightteaserwidth}{-3em}
\captionsetup{labelfont=bf,textfont=it}

\teaser{%
\begin{subfigure}[t]{0.37\textwidth}
  \includegraphics[height=5.4cm]{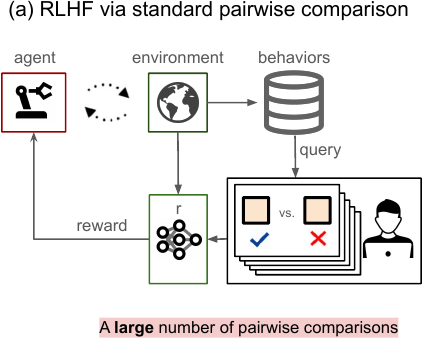}
  \caption{Standard RLHF requires high workload due to pairwise comparisons, and provides no ability to steer the process.}
  \label{fig:teaser-a}
\end{subfigure}%
\hspace{1em}%
\begin{subfigure}[t]{0.6\textwidth}
  \includegraphics[height=5.4cm]{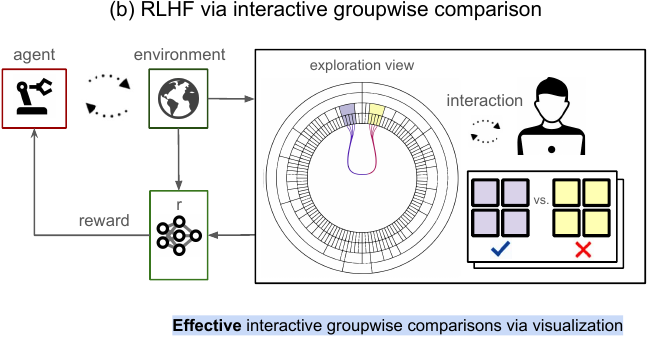}
  \caption{Our RLHF approach requires less work due to groupwise comparisons, and the user can steer the process actively.}
  \label{fig:teaser-b}
\end{subfigure}%
\caption{The standard RLHF uses pairwise comparisons and therefore requires a large number of comparisons leading to a high workload. The comparison pairs are suggested by the system and cannot be chosen by the user. Our RLHF approach provides more agency to the user and demands less work: we leverage the user's visual abilities to effectively explore the behaviour space via hierarchical grouping in the `exploration view' and to select groups for comparison.}%
\label{fig:teaser}%
}

\maketitle
\begin{abstract}
 Reinforcement learning from human feedback (RLHF) has emerged as a key enabling technology for aligning AI behaviour with human preferences. The traditional way to collect data in RLHF is via pairwise comparisons: human raters are asked to indicate which one of two samples they prefer. We present an interactive visualisation that better exploits the human visual ability to compare and explore whole groups of samples. The interface is comprised of two linked views: 1) an exploration view showing a contextual overview of all sampled behaviours organised in a hierarchical clustering structure; and 2) a comparison view displaying two selected groups of behaviours for user queries. Users can efficiently explore large sets of behaviours by iterating between these two views. Additionally, we devised an active learning approach suggesting groups for comparison. As shown by our evaluation \alladded in six simulated robotics tasks, our approach increases the final rewards by 69.34\%\allsame. It leads to lower error rates and better policies. We open-source the code that can be easily integrated into the RLHF training loop, supporting research on human-AI alignment.
\end{abstract}  


\section{Introduction}
\begin{SCfigure*}
\begin{tikzpicture}%

\node (A) at (0,0) [inner sep=0, anchor=south west] {\includegraphics[width=0.6\textwidth]{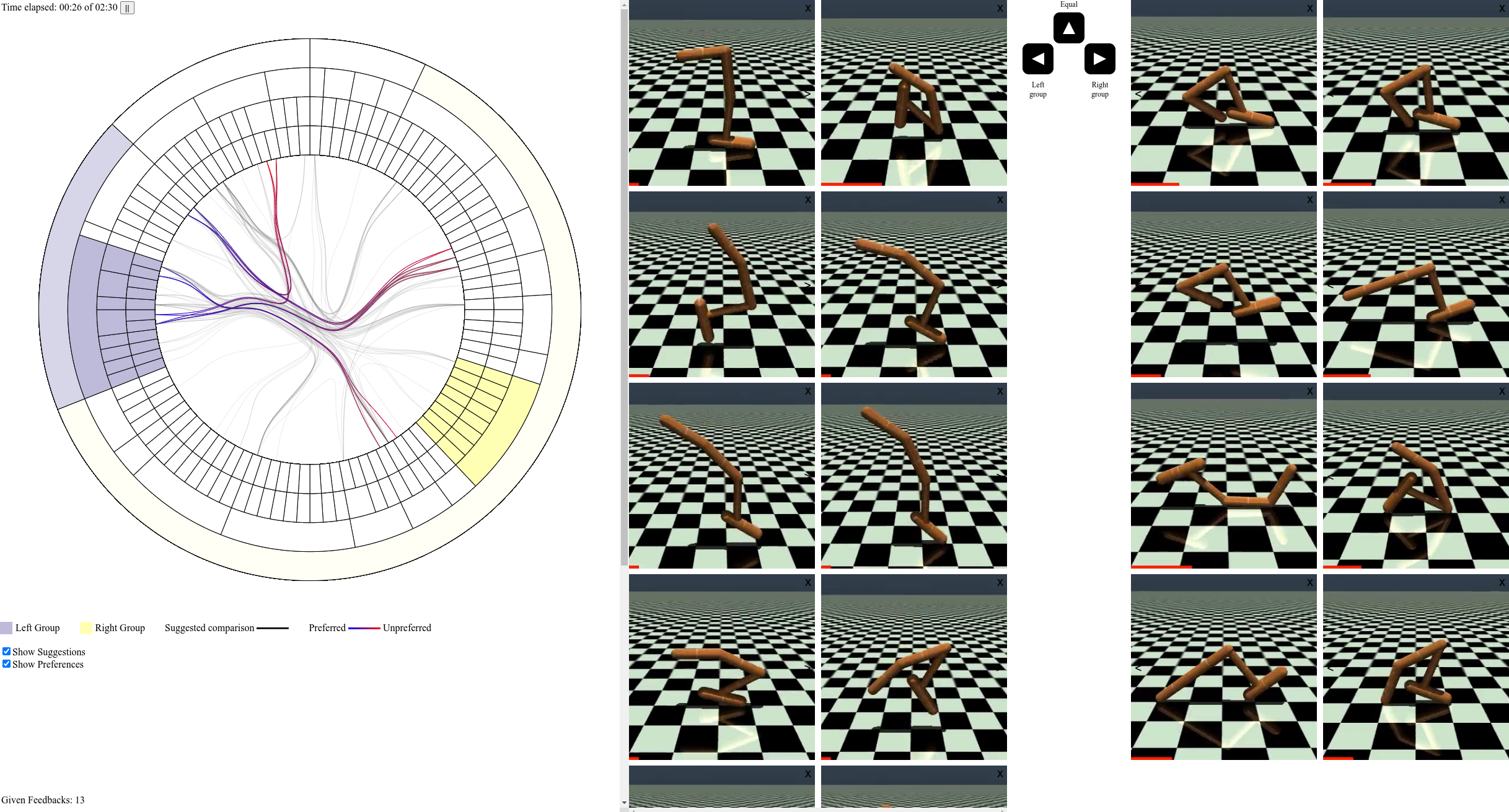}};

\fill [white] (0, 0) rectangle (4, 1.5);
\fill [white] (0, 5.5) rectangle (1, 6);


\coordinate (E) at (2.25, 6);

\begin{scope}[font=\itshape]
\node at (E) [anchor=south] {exploration view};
\end{scope}

\coordinate (C) at (7.6, 6);

\begin{scope}[font=\itshape]
\node at (C) [anchor=south] {comparison view};
\end{scope}

\begin{scope}[font=\tiny]
\node (IB) at (0, 5.4) [anchor=south west, align=left] {individual\\behaviors};
\node (GB) at (4.2, 5.4) [anchor=south east, align=right] {groups of\\behaviors};
\node (PA) at (0, 1) [anchor=south west, align=left] {previously\\compared behaviors};
\node (SC) at (4.2, 1) [anchor=south east, align=right] {suggested\\comparisons};
\node (YES) at (PA.south west) [yshift=-0.5\baselineskip, anchor=base west] {yes};
\node (NO) at (PA.south east) [yshift=-0.5\baselineskip, anchor=base east] {no};

\node (G1) at (SC.south west) [yshift=-0.5\baselineskip, anchor=base west] {};
\node (G2) at (SC.south east) [yshift=-0.5\baselineskip, anchor=base east] {};
\end{scope}

\shade [left color=blue, right color=red] ($(YES.base east) + (0, 0.0)$) rectangle ($(NO.north west) + (0, -0.1)$);
\fill [gray] ($(G1.base east) + (0, 0.0)$) rectangle ($(G2.north west) + (0, 0.0)$);

\coordinate (IB1) at (1.05, 3.87);
\coordinate (IB2) at (1.24, 4.265);
\coordinate (IB3) at (1.52, 4.53);

\coordinate (GB1) at (3.2, 4.8);
\coordinate (GB2) at (3.55, 4.4);
\coordinate (GB3) at (3.95, 4);

\coordinate (TPA) at (1.6, 3.5);
\coordinate (TSC) at (2.9, 3.525);

\begin{scope}[every edge/.append style = { ->, >=stealth, black}]

\draw [bend right=10] ($(IB.south) + (-0.025,0)$) edge (IB1);
\draw [bend right=10] (IB.south) edge (IB2);
\draw [bend right=10] ($(IB.south) + (0.025,0)$) edge (IB3);

\draw [bend left=10] ($(GB.south) + (-0.025,0)$) edge (GB1);
\draw [bend left=10] (GB.south) edge (GB2);
\draw [bend left=10] ($(GB.south) + (0.025,0)$) edge (GB3);

\draw [bend right=10] (PA.north) edge (TPA);
\draw [bend left=10] (SC.north) edge (TSC);

\end{scope}

\end{tikzpicture}
\caption{Our user interface's two connected views:
  On the left, the \emph{Exploration View} 
displays the model-supplied sampled behaviors
in a hierarchically arranged radial chart.
Users can select groups or individual behaviors for comparison, using a mouse.
Gray lines denote machine suggestions for comparisons,
while the system presents previously made comparisons in color.
This view shows two groups of videos,
and the user's task is to specify a preference -- i.e.,
to state which group comes closer to the behavior desired from the agent.
Users can edit these groups at any time by adding or deleting videos,
or even moving them from one group to the other.}%
\label{fig:ui}%
\end{SCfigure*}

\alladded
Over the last decade, success with \allsame \emph{reinforcement learning} (RL) has expanded 
to a wide range of difficult tasks, among them Go \cite{Silver2016MasteringTG}, Dota 2 \cite{Berner2019Dota2W}, and Atari games \cite{Mnih2015HumanlevelCT}. The main idea in this method of experience-based training for artificial-intelligence (AI) models \cite{sutton2018reinforcement}
is to reward preferred behaviours 
and punish the non-preferred ones. \changed{By optimising for reward (that is, a numerical value from a \emph{reward function} that judges the model's behaviour), the AI model improves its \emph{policy}, a mapping from each state to an action}.
In this context, \emph{behaviours} are segments of the agent's state--action sequences. For example, one might be part of a written response; a generated image; or a time series of angles, positions, and torques of a robot's joints. \changed{The \emph{behaviour space} encompasses the set of all behaviours sampled as of the given time to present for human inspection.}

Reward functions pose a problem, however. It has 
proven highly challenging to link them to 
human preferences by means of mathematical equations
\cite{mahadevan1996average,gupta2024behavior}.
Therefore,
research has turned to
human feedback as a source of guidance
for RL.
One of the most popular methods of this kind is \emph{reinforcement learning from human feedback} (RLHF)~\cite{Christiano2017DeepRL}.
It applies the principle that, 
repeatedly, human evaluators shown two behaviours
by an AI model (e.g., distinct images/videos) 
choose
which of the two they prefer, thus generating
preference data
that function in training a \emph{reward model},
which serves as the reward function
~\cite{Christiano2017DeepRL, ziegler2019fine, ouyang2022training}.
This concept has informed work on
such hard-to-formulate objectives
as
safety \cite{Dai2023SafeRS},
factuality \cite{Sun2023AligningLM}, 
and aesthetics \cite{Wallace2023DiffusionMA}; on
fine-tuning models to generate images better aligned with human preferences \cite{Black2023TrainingDM,Lee2023AligningTM}; and on stronger grounding for textual explanations of images \cite{Yu2023RLHFVTT}.

Standard RLHF~\cite{Christiano2017DeepRL}
relies on \emph{pairwise comparisons} -- 
that is,
asking users to judge many pairs of behaviours in succession
(see Fig.~\ref{fig:teaser}-a)
until a set comparison-count threshold is reached.
This is highly labourious work, though,
and leaves no agency for the users,
who cannot choose which behaviours to compare.
It displays major limitations:
\begin{itemize}
    \item \emph{Time-inefficiency:} Looking at one pair of behaviours at a time is time-consuming. Gathering enough human feedback requires 700-plus comparisons for even simple behaviours such as a robot walking forward~\cite{Christiano2017DeepRL}. It is labour-intensive and costly.
    \item \emph{Lack of user agency:} Users have a clear idea
of what the desired agent behaviour should look like, yet standard RLHF does not permit them to explore and select behaviours \changed{\emph{interactively}}, to provide more effective feedback. 
    \item \emph{Inability to exploit context-related information:}
    Standard implementations neither show nor utilise valuable context information:
    there is no \changed{abstract} overview of all behaviours,
    list of the comparisons already made by the user, and so forth
    Users are left unable to understand the broader behaviour space.
\end{itemize}
Therefore, standard RLHF can be impractical in applications that
require the expertise of a specialist 
(e.g., a medical doctor)
and in settings wherein the aim is to teach a model for a creative purpose
(such as game design).
Such cases would
benefit from
more user agency
and
a lower workload
than the standard RLHF approach affords \cite{danielskoch2022expertiseproblemlearningspecialized}.

We set out to
expand the reward-elicitation interface substantially
by visualising the entire behaviour space
in a hierarchical manner
and letting users navigate it freely, thereby 
exercising greater agency.
We enabled them to
categorise behaviours
into separate groups,
then compare these groups with each other.
Our tests showed that this increases efficiency; 
more preferences can be recorded in a given time span than with the standard approach\changed{, and the final policy returns are improved, thanks to letting the user give more informative feedback}.
Furthermore, we decisively augmented the behaviour space's depiction
by visualising the progress thus far
and suggesting sets to compare next.
The user gains power to review earlier decisions
and adapt the work accordingly.
Figure~\ref{fig:ui} illustrates the interface created.

Prior work
tackling user interfaces
for RLHF
addresses
modular environments'
role in rapidly developing such interfaces
and analysing the feedback \cite{Metz2023RLHFBlenderAC, Yuan2024UniRLHFUP}.
Groupwise comparison
of behaviours
has been discussed especially
by Zhang et al.~\cite{zhang2022time}.
Our work goes significantly further, as Section~\ref{sec:background_visui} details, by advancing
how the behaviour space is visualised \changed{(via a hierarchical structure with information on the feedback history instead of a scatterplot)},
the complexity of the RL cases addressed \changed{(with movement patterns rather than still poses as goal behaviours)},
and the level of empirical evaluation \changed{(using synthetic studies in six environments, based on those environments' original reward functions, plus a non-synthetic one -- as opposed to only synthetic evaluation in three environments with simplified reward functions)}.

This paper reports on several vital contributions:
\begin{itemize}

\item A novel user interface for \changed{interactive }groupwise preference elicitation for RLHF based on thorough data and task abstractions (see Sec.~\ref{sec:system})

\item Three case studies centred on \changed{interactive }groupwise comparison
in training for complex novel behaviours in the robotics domain (see Sec.~\ref{sec:usageexamples})

\item A simulation study demonstrating that the interactive groupwise approach outperforms the standard pairwise one \changed{by 69.34\% in its policy return} (see Sec.~\ref{sec:simulationstudy})

\item Evaluation with experts that identified a marked increase in efficiency: eliciting 86.7\% more preferences with interactive groupwise comparison than with standard pairwise comparison (see Sec.~\ref{chap:userstudy})

\item Open-source code to support further research at ~\url{ https://jankomp.github.io/interactive\_rlhf}.

\end{itemize}

\section{Background and Related Work}
For a backdrop to our work, this section lays out the key concepts behind RLHF in overview, then reviews work that informed our project: research into visualisations for RL and corresponding behaviour data.

\subsection{Prerequisites for RLHF.}
Reinforcement learning, the branch of machine learning that teaches autonomous agents how to perform tasks by interacting within an environment~\cite{sutton2018reinforcement}, 
is formally defined as the sequential decision-making problem expressed by the Markov decision process (MDP) $ \mathcal{M} =\langle \mathcal{S}, \mathcal{A}, \mathcal{T}, R, \gamma \rangle$.
At each time step $t$, the agent receives the state observation $s_t \in \mathcal{S}$ from the environment, where $\mathcal{S}$ is the set of possible states. The agent interacts with the environment by performing action $a_t$ from the action space, $\mathcal{A}$. The environment enters the
the next state, $s_{t+1}$, as defined by the state--action transition function, $\mathcal{T}$. At each step, the agent receives a scalar reward ($r$) from the reward function ($R$) that reflects the agent's performance in pursuit of the goal set. An RL agent learns to maximise the accumulated reward through a trial-and-error process by trying out actions and observing the resulting reward, which it tries to maximise.
Usually, the policy training employs deep reinforcement learning~\cite{franccois2018introduction}, which combines traditional RL algorithms with deep neural networks. This enables their use with potentially complex material, such as images and other high-dimensionality observations.

In formulating an MDP for a real-world application, designing the reward is perhaps the most decisive \cite{silver2021reward} but, at the same time, most challenging element~\cite{mahadevan1996average,gupta2024behavior}.
Empirical evidence demonstrates the effectiveness of incorporating human preferences into RL to enhance robotics~\cite{abramson2022improving, hwang2024promptable} and to tune large language models (LLMs)~\cite{ziegler2019fine, ouyang2022training}.
Learning from user priorities exhibits greater efficiency when users compare state--action trajectories by expressing their preferences~\cite{wirth2017survey, akrour2012april, furnkranz2012preference}, relative to when from users demonstrate their desires~\cite{ng2000algorithms, abbeel2004apprenticeship}. 
Research in the RL field has homed in on various methods of incorporating human feedback, with rankings, ratings scales, clustering, decomposition, and so forth~\cite{akrour2011preference, pilarski2011online, akrour2012april, daniel2015active, el2016score, zintgraf2018ordered,shi2025dxhf}.

Meanwhile, other research has explored using preferences rather than absolute rewards for reinforcement learning~\cite{furnkranz2012preference, akrour2014programming}.
Christiano and colleagues studied how to elicit human preferences from pairwise comparisons of trajectory segments~\cite{Christiano2017DeepRL}. 
This process of reward-modeling from human feedback involves learning a user's preferences from among several options by collecting feedback from that user. Users are asked to indicate their preferences via relative feedback, such as ``I prefer A over B.'' Often, the preference elicitation is based on pairwise comparison, where a user query is defined as $q = \{ ({\tau_i}, {\tau_j}; \operatorname{o}) \}$, with
$\operatorname{o} = \{\prec, \succ, \sim \}$ indicating the preference relation between the two trajectories. The preference order is commonly defined on the basis of estimated expected return $\hat{R}$ for the trajectories $\tau_i$ and $\tau_j$.
\allchanged
The following equation captures how the user decides on the preference order in light of the expected return implicitly given for each trajectory and the level of noise:
\begin{equation}
\label{eq:pref_order}
\operatorname{o}(\tau_i, \tau_j) =
\begin{cases}
\tau_i \succ \tau_j & \text{if } \hat{R}(\tau_i) + \epsilon_i > \hat{R}(\tau_j) + \epsilon_j \\
\tau_i \prec \tau_j & \text{if } \hat{R}(\tau_i) + \epsilon_i < \hat{R}(\tau_j) + \epsilon_j \\
\tau_i \sim \tau_j & \text{if } \hat{R}(\tau_i) + \epsilon_i = \hat{R}(\tau_j) + \epsilon_j \\
\end{cases}
\end{equation}

\allsame

where $\epsilon$ is a term specifying the level of random noise \changed{that affects the perception of an instance}. The average magnitude of $\epsilon$ is influenced by factors such as
the annotator's cognitive abilities; in presentation of a behaviour for the relevant query, this noise affects how the user internally assesses the expected return for the behaviour, $\hat{R}(\cdot)$, 
\allchanged and chooses accordingly.
Through training on the order of preference from many pairs, a neural network can be trained to predict the expected return of a trajectory directly, by means of the Bradley--Terry model \cite{bradley_terry}. Loosely characterised, the neural network serves as the inverse of the function \ref{eq:pref_order}; that is, the model learns to predict $\hat{R}(\cdot)$ for a trajectory $\tau$ from a set of $n$ queries $\mathcal{Q} = \{q_1, ..., q_n\}$. \allsame

\subsection{visualisations and Graphical Interfaces for Deep RL.}
\label{sec:background_visui}
\allchanged
Liu et al. demonstrated the power of visual analytics to enhance explainability and aid in implementing explainable AI \cite{liu2024vizforai}. Other researchers created the tool SampleViz, to help visually with analytics for RL and in debugging \cite{liang2024sample}.
Scholarship is concerned also with visual analytics tools in the setting of multi-Agent RL \cite{shi2023marlviz,zhang2024MARLens}. Visual analytics has shown use not only with RL but also for new machine-learning paradigms such as foundation models~\cite{yang2024foundation}.
\allsame

Along similar lines, 
DQNViz~\cite{wang2018dqnviz} offers a multi-level visualisation system for Deep Q-Networks, incorporating training statistics, trajectory displays, and segment-level details. This system enables users to diagnose agent behaviours and refine strategies by means of interactive visual exploration of agent experiences in Atari environments.
With focus on recurrent-neural-network-based deep-RL agents, DRLViz~\cite{jaunet2020drlviz} and DRLIVE~\cite{wang2021visual}, in turn, visualise their internal memory representations. 
Recently, the VISITOR framework~\cite{metz2023visitor} has expanded on these approaches, as a general way of exploring state sequences. 
Efforts with Interactive Reward Tuning~\cite{shi2024interactive}, RLHF-Blender~\cite{Metz2023RLHFBlenderAC}, and DxHF~\cite{shi2025dxhf} emphasise human-in-the-loop approaches for AI alignment, letting users interactively modify reward functions or provide multiple types of feedback for reward modeling. 

The closest relative to our technique is CLRVis, by Zhang et al.~\cite{zhang2022time}, alluded to above.
They too considered groupwise comparisons of behaviours.
In their system, human labellers explore the behaviour space t-SNE projections.
Yet t-SNE is not ideal for such a scenario, as we show in Subsec.~\ref{sec:cluster},
since this mechanism often fails to group similar behaviours together.
\changed{\st{Hence, we pursued for a hierarchical approach that outperforms t-SNE.}}
Furthermore,
the CLRVis system creates the dataset from a comparison of two groups (of sizes $m$ and $n$) by enumerating all possible pairs and hence obtaining
$m \times n$ pairwise comparisons -- a method
not evaluated by real users.
We visualise the behaviour space quite differently,
create the training dataset differently, and
also show comparison progress and suggestions.
In further contrasts,
CLRVis focuses on ranking time steps (images),
whereas we enable ranking of sequences (videos).
Also, we conducted empirical evaluation of our setup
and can attest to its utility for more complex tasks.

\alladded
\subsection{Visual Interactive Labeling.}
Visual interactive labelling (VIL) constitutes another area of related work. 
Several studies by Bernard et al. \cite{bernardTowards,bernardVIAL,bernardTaxonomy} address this technique, in which the users find the samples in need of labels with the aid of data visualisations, whereas active learning (AL) relies on algorithms to find those samples (in both cases, the user does the labelling). From comparing the two especially \cite{bernardComparing}, their findings underline that VIL can outperform AL provided that the dimensionality-reduction technique separates the data well.
Crucially, VIL can help bridge the `cold start' problem that plagues AL. Therefore, the team provided a strong foundation on which we could build, even though they did not work on RLHF or examine VIL's utility in the context of providing human feedback for iteratively training an RL agent.
In other work, Matt et al.~\cite{MATT2025104240} highlighted the advantage of shifting from class-to-instance assignment to instance-to-class assignment in VIL. Our work differs in focusing on preference ratings for RL rather than classification.  
Also, our interface-design goals were informed by Bernard et al.'s proposal of an optimal labelling strategy that consists of a `Discovery, Consolidation, and Fine-Tuning' phase~\cite{10.5555/3290776.3290797}.
Finally, Grossmann et al.~\cite{grossmann2021doeslayoutreallymatter} showed that interface layout has little effect on accuracy estimation in VIL. Therefore, rather than seek the single most effective layout, we focus on enabling interactions that support an optimal labelling strategy.

\allsame

\subsection{Visualising Agent Behaviours.}
One can regard behaviours of RL agents as event sequences of varying length~\cite{wu2020discovering}.
Hierarchy-based visual representations adeptly organise and aggregate these sequences, affording the emergence of valuable insight.
For instance, LifeFlow~\cite{wongsuphasawat2011lifeflow} exploits a tree structure to visualise common patterns through icicle plots (which depict hierarchical data by using rectangular sectors that cascade from root to leaves \cite{59067945-6bf5-3b4f-b219-127b3de0516c}).
Similarly, CoreFlow~\cite{liu2017coreflow} illustrates branching patterns via nodes and links to highlight frequently occurring paths.
GestureAnalyzer~\cite{jang2014gestureanalyzer}, in turn, portrays hierarchical clustering of behaviours in a pose tree, visually representing motion trends. The follow-on system MotionFlow~\cite{jang2015motionflow} emphasises transitions' depiction through flow diagrams and facilitates direct user interaction to refine a `pose tree'.
Other progress has come from ViewFusion~\cite{trumper2012viewfusion} which combines hierarchical structures with time-dependent activities by using treemaps, and from
ActiviTree~\cite{vrotsou2009activitree} which offers an interactive node--link tree layout for exploring event sequences.
Although these methods capture hierarchical relationships well, displaying the data linearly can limit inter-group comparisons. We sought a design that enriches the hierarchical relations' graphical presentation, facilitates clustering, and improves comparative analysis.

\section{Interactive Groupwise Comparison of behaviours}
\label{sec:system}
To describe our approach, we begin by characterising the data
and
the task analysis undertaken,
then, from the starting point thus provided, describe
the approach
to rendering
the behaviour space visually
accessible to users
through data clustering and \changed{discuss three design alternatives}
and the visualisation interface.
Integrating these methods
into an interactive RLHF system
culminated in the novel approach
of interactive groupwise comparison
for RLHF.

\subsection{Characterisation of the Data.}
\label{sec:dataabstraction}
The behaviour space comprises the sequences of states and agent actions.
What a state represents depends on the agent,
but
a state generally can be understood as a vector of numbers,
with the agent's behaviour being a series of these vectors.
In our robotics examples (see Sec.~\ref{sec:usageexamples}),
the joint angles and positions
of the robotic skeleton constitute the state,
which could easily extend to
20 dimensions or more.

Our work focused on agents
whose behaviour can be represented
in the form of videos or images\changed{\st{ such that the user
can rate them}}.
Although the system was not set up to learn from the videos or images themselves, doing so would not fundamentally change the setup,
as long as a distance metric between videos can serve
to represent each behaviour in a lower-dimensional space.
Since the state changes over time,
it can be handled
as $n$-dimensional
time-series data.
behaviours
are likely to
differ in their length,
so all analysis and visualisation algorithms
applied had to be able to cope with multi-variate time-series data of varying lengths.
User exploration of the behaviour data
(detailed in \ref{sec:taskanalysis})
created a need for reducing the dimensionality of the behaviour space.

\changed{For our purposes, a sequence's frame rate is 30 images per second. Since the video clips in this setting are one second long, each behaviour is defined as a sequence of 30 frames. Other settings should be equally viable, though. Longer clips can give the model more information but grow harder for a human to judge. This tradeoff needs to be balanced.}

\subsection{Task Description.}
\label{sec:taskanalysis}
In a round of feedback in RLHF, users are asked to \emph{identify desirable and undesired behaviours}
when presented with behaviours of an agent
(typically, 100--200 \allchanged~\cite{Christiano2017DeepRL,gleave2022imitation}\allsame).
On this basis, the behaviours identified as better get rewarded
such that the system can learn from human feedback.

The standard RLHF approach
boils this overarching task
down to a beguilingly simple one: \emph{pairwise comparisons} (see Fig.~\ref{fig:teaser}-a).
The user is shown two behaviours, \changed{$\tau_i$} and \changed{$\tau_j$},
in the form of an image or video.
Then the user has to state a
preference, for \changed{$\tau_i$} over \changed{$\tau_j$}
or the other way around.
It is possible also to specify no preference
between \changed{$\tau_i$} and \changed{$\tau_j$}.
The pairwise-comparison task
is easy for humans to understand
but brings with it a massive workload,
since updating the underlying model
requires quite a large number of such comparisons.
Furthermore,
without agency over which behaviours to compare,
the user ends up acting as a mere tool
performing a simple task many times, stoically.

We strove, then, for a system that gives the user more agency in the interaction.
For this we still addressed the overarching task defined above,
but we broke it into four sub-tasks.
\begin{enumerate}[labelindent=0pt, labelwidth=\widthof{\textbf{\ref{task:progress}}},itemindent=1em,leftmargin=!,label=\textbf{T.\arabic*}, ref=T.\arabic*]
\item \label{task:find}: \emph{Explore behaviours from among the set of all behaviours}
\item \label{task:categorize}: \emph{Class behaviours into two groups (preferred versus non-preferred)}
\item \label{task:compare}: \emph{Compare between groups of behaviours}
\item \label{task:progress}: \emph{Track the progress of comparison}
\end{enumerate}
This approach calls for more agency on users' part,
since they can see which behaviours have / have not been compared and are empowered to select the behaviours to compare next.
Machine support remains available
but as an offer that need not be accepted.
Furthermore, we capitalise
on humans' natural pattern-recognition:
people can categorise behaviours and group them together~\cite{zelazo2013oxford}.
Lastly,
groupwise comparison
leads to a much lower workload\changed{~\cite{zhang2022time}},
thanks to working with more feedback at a time, than pairwise comparison (see Fig.~\ref{fig:teaser}-b).


\allchanged
\subsection{Hierarchical Structuring of the behaviour Space.}
\label{sec:cluster}

Users engaging with RLHF need support for exploring given behaviours, categorising them into preference groups, comparing groups, and tracking progress. To make the behaviour space interpretable, we evaluated three methods: dimensionality reduction via principal component analysis (PCA)~\cite{mackiewicz1993principal}, t-SNE~\cite{van2008visualizing}, and agglomerative hierarchical clustering~\cite{day1984efficient}. All operated on pairwise distances computed with dynamic time warping~\cite{salvador2007toward}.

Dimensionality-reduction methods yield a 2D embedding that can be visualised as a scatterplot. However, group-selection then requires an additional clustering step performed by the user. In contrast, hierarchical clustering directly produces clusters at multiple levels of granularity~\cite{cohenaddad2017hierarchicalclusteringobjectivefunctions}, enabling immediate selection of groups through the tree structure.

To compare methods, we measured intra-cluster variance with respect to the true reward in 10 runs for an environment for which the true reward is known~\cite{towers2024gymnasium, erez2012infinite}. Paired $t$-tests showed that hierarchical clustering (hc) yields significantly lower intra-cluster variance than PCA or t-SNE (hc $<$ PCA: $t=6.54, p=0.0001$; hc $<$ t-SNE: $t=4.47, p=0.0016$). We concluded that hc forms more reward-consistent groups.

In summary, we found hierarchical clustering better suited to our purpose: it avoids an additional grouping step and supports multi-level exploration. The supplementary material supplies details of the data-processing pipeline and documents our analysis of the design \cite{komp2025interactive}.

\allsame

\subsection{visualisation of the behaviour Space.}
\label{sec:vis}
\changed{For a solid design, it is crucial that the user not rely solely on trial and error when selecting behaviours. Therefore, the exploration view incorporated cues that the user can take as guidance.}
In addition
to the hierarchical relations between behaviours,
the data depiction captures
adjacency relations between the leaves
in the hierarchy --
namely, which behaviours the user has compared with each other
and which behaviours the system recommends for comparison.
This directly supports
handling sub-task~\ref{task:progress}.
%
Following Holten's lead in detailed exploration of graphically
presenting hierarchies with adjacency information~\cite{Holten2006HierarchicalEB},
we experimented with several options \changed{for visualising adjacency relations} 
(the most promising are shown in \changed{\st{Fig. 3}the supporting information~\cite{komp2025interactive}).}
Proceeding from these experiments
alongside Holten's work and that
of Schulz \cite{schulz_treevis_2011}, we reached the following conclusions with regard to \ref{task:find}, \ref{task:categorize}, and \ref{task:progress}:
%
%
\begin{itemize}
\item
The layout needs to be \emph{space-efficient}.
Treemaps and radial layouts
make good use of the space available
and can accommodate quite a few nodes on the screen,
whereas classical node--link representations are likely to waste space
from growing more in one direction than the other.

\item
Each node in the tree must be \emph{selectable} by the user.
Radial layouts and node--link representations simplify this
while treemaps produce item overlaps that complicate it.

\item
The selectable nodes should remain \emph{non-obstructed} by the adjacency information.
Although treemaps' space-filling nature precludes that,
a few radial layouts can meet this requirement.
Neither classical node--link nor icicle representations
excel here~\cite{Holten2006HierarchicalEB}.

\end{itemize}
%
We decided on the radial layout
at the left in Figure~\ref{fig:ui}. In such a layout,
the innermost segments represent leaf nodes
and the outer layers point to parent nodes.

Within the radial chart,
curved lines connecting leaf nodes
express relations between behaviours.
There are two types of lines:
gray lines denote comparison recommendations \allchanged based on the variance in the predictions\allsame,
while coloured ones track the user's feedback history,
maintaining guidance and orientation throughout the interaction.
The lines are bundled together to avoid visual clutter. 
\allchanged
The main purpose behind the gray lines is to give a window to which comparisons the reward predictors are most ``unsure'' about. While the coloured lines' core purpose is to record which comparisons have been made, so that these are not repeated, they fill a secondary function of supporting exploration by presenting the preferred versus non-preferred behaviours from each past comparison. Although endpoints in different colours may connect similar behaviours, users can easily track their past decisions, both avoiding repetition and benefiting from the colour gradient as a cue for remembering their preferences.

To prevent undue influence on the user's decision, the interface omits the absolute predictions of the reward and the mean of the predicted rewards. For RLHF, it is pivotal that users focus on their own preference without being swayed by external inputs. \allsame

\subsection{Interactive Comparison of behaviour Groups.}
\label{sec:learn}
At the heart of the user-interface design are two views:
dubbed the \emph{Exploration View} and the \emph{Comparison View} 
(see Fig.~\ref{fig:ui}). 
\changed{The behaviours visualised in these views are linked.}
Clicking any item during exploration
selects the corresponding behaviour set
for comparison.
Importantly, clicking a parent node brings in all behaviours falling under it,
and individual behaviours can be added to or removed from a group by clicking on the relevant leaf.
This combines efficiency with free selection of behaviours.
From the view on the right, which displays the groups selected for comparison,
the user can judge between two groups of behaviours (per~\ref{task:compare}),
deciding which group matches the desired outcome more.
Managing each group by removing outliers or transferring behaviours to the other group before the choice 
enhances flexibility and the preference feedback's quality both. 
After providing a preference decision, users are free to select the next groups via the Exploration View or request the system-recommended group-comparison query.
\allchanged

\emph{Label generation}:
\allsame
Given the feedback on the two groups, of sizes $m$ and $n$, the system samples $\max(m,n)$ pairs of behaviours from these groups as the preference data. Instead of using the Cartesian product of the two groups for the sampling pool
(giving us $m \times n$ pairs)\changed{, we sample $\max(m,n)$ pairs such that each group member is present in a pair at least once}. Sampling fewer pairs per group comparison avoids overfitting to a suboptimal reward.
When we \changed{sampled with the Cartesian product} as proposed by Zhang et al. \cite{zhang2022time}, the resulting policies often got stuck in local optima.
Hence, we turned to $\max(m,n)$ pairs of behaviours.

\allchanged
\emph{Active learning for suggestions}:
\label{par:group_var_score}
Our approach adapts techniques from active-learning heuristics in pairwise comparison \cite{Christiano2017DeepRL} to groupwise comparisons by calculating a \emph{group-variance score}. This score is higher if the reward predictors are uncertain which of the two groups is the better one (finding the correct choice requires querying a human) and is lower if they are less certain about the quality of the behaviours within the groups (that is, when the groups are not uniform). Also, it applies a slight penalty for size differences between the groups. For details, refer to the supporting information \cite{komp2025interactive}.

\subsection{Visual Mapping of Tasks}
In summary, \ref{task:find} is mapped to a hierarchical radial chart (HRC) within the exploration view. \st{Showing a hierarchical structure with the behaviours ordered radially, in groups, with the most similar behaviours nearby, equips users to explore the behaviour space.} Sub-task \ref{task:compare} maps to the comparison view. \st{The user can assess the behaviours there by inspecting the behaviours themselves and rate them by means of the buttons or the arrow keys.} The mapping for \ref{task:categorize} uses the two views in combination. \st{Users use the HRC for the selection and receive immediate feedback in the comparison view about what they selected in the comparison view. Then, they refine or change their selections by clicking the HRC or comparison view respectively.} Finally, \ref{task:progress} maps to employing the HRC's edge bundles as part of the exploration view and displaying the number of feedback instances completed.
\allsame

%

\section{Example Cases}
\label{sec:usageexamples}
Deployments in three MuJoCo \cite{todorov2012mujoco} environments commonly utilised in RL and RLHF research~\cite{Christiano2017DeepRL} (HalfCheetah, Walker, and Hopper) demonstrate our approach. MuJoCo is a popular free and open-source physics engine designed to facilitate research and development in robotics, biomechanics, graphics, animation, and other areas necessitating fast yet accurate simulation~\cite{todorov2012mujoco}. \changed{Although we follow prior
work's lead in placing the robotics domain in an illustrative role, our approach holds potential for video games, image- and video-generation, and many other applications.}
The objective is to arrive at policies to execute behaviours for which no predefined reward functions exist; \changed{RLHF can help us train such policies. Using the interface affords purposeful provision of comparisons linked to the final intended behaviour}.

\newlength{\thispicwidth}
\setlength{\thispicwidth}{0.1\textwidth}

\begin{figure}
\centering
\begin{subfigure}[b]{0.5\textwidth}%
\includegraphics[width=\thispicwidth]{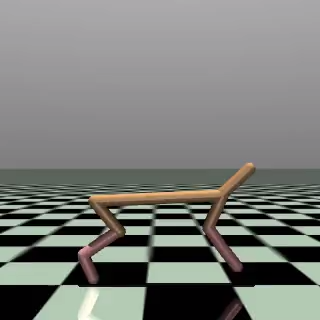}
\includegraphics[width=\thispicwidth]{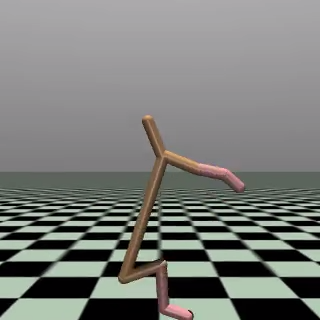}
\includegraphics[width=\thispicwidth]{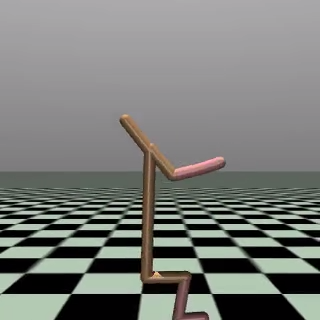}
\includegraphics[width=\thispicwidth]{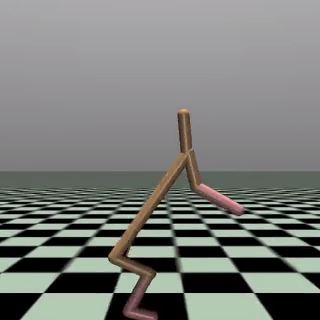}
\includegraphics[width=\thispicwidth]{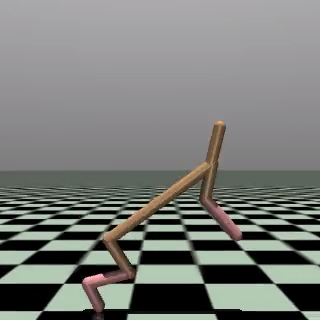}%
\caption{HalfCheetah learned to stand up from 601 binary preferences.}
\end{subfigure}%
\\%
\begin{subfigure}[b]{0.5\textwidth}%
\includegraphics[width=\thispicwidth]{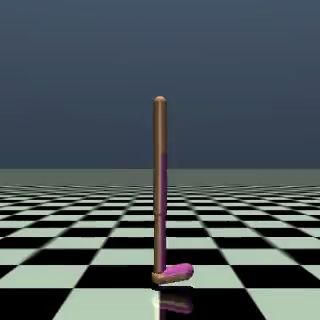}
\includegraphics[width=\thispicwidth]{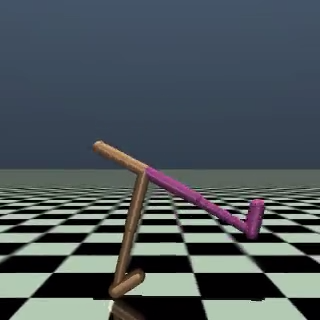}
\includegraphics[width=\thispicwidth]{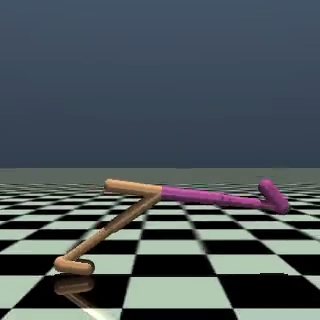}
\includegraphics[width=\thispicwidth]{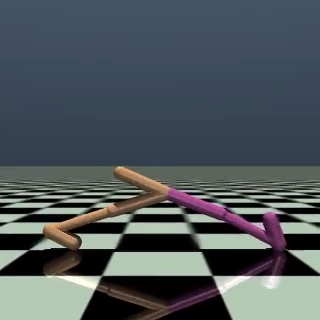}
\includegraphics[width=\thispicwidth]{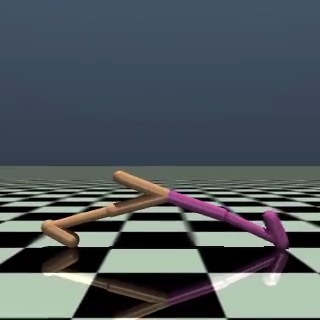}%
\caption{Walker learned to do the splits from 616 binary preferences.}%
\end{subfigure}%
\\%
\begin{subfigure}[b]{0.5\textwidth}%
\includegraphics[width=\thispicwidth]{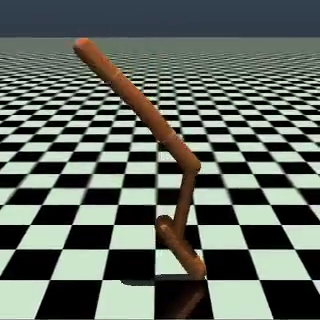}
\includegraphics[width=\thispicwidth]{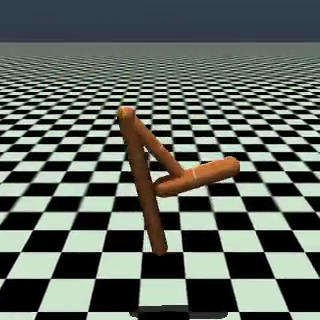}
\includegraphics[width=\thispicwidth]{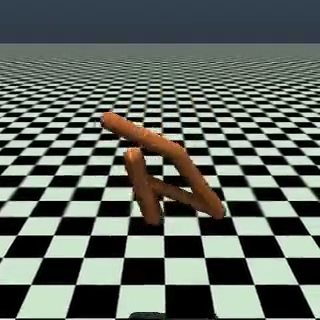}
\includegraphics[width=\thispicwidth]{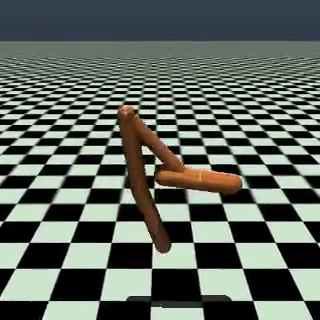}
\includegraphics[width=\thispicwidth]{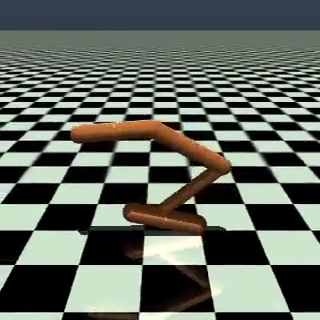}%
\caption{Hopper learned to double-backflip from 1,032 binary preferences.}%
\end{subfigure}%
\vspace{-2.0ex}%
\caption{
behaviours not coupled with ground-truth rewards can be effectively learned from interactive groupwise comparisons.
For each behaviour, we present a five-frame sequence.}
\label{fig:cases}
\vspace{-4.0ex}
\end{figure}
\allchanged

\allsame

The first of the three interactive-groupwise-RLHF case studies used \textbf{HalfCheetah}. Its original objective was to apply torque to the joints for running in the forward direction.
After approximately 35 min of \changed{\st{interaction with the interface}exploration, selection of groups, and groupwise comparison via our interface, we had} 601 preferences, since each comparison led to
multiple preferences. HalfCheetah succeeded in learning how to stand up and sit.
Secondly, we taught \textbf{Walker}, a robot originally designed to walk forward by means of torques to the six hinges connecting its seven body parts, to do the splits, from 616 preferences \changed{found in 40 min of exploring, selecting groups, and making groupwise comparisons}. 
Finally, the initial goal with \textbf{Hopper} was to move the hopper forward in hops by applying torques to the three hinges connecting its four body parts. Drawing inspiration from Christiano et al.~\cite{Christiano2017DeepRL}, who taught it to do backflips with RLHF from 900 pairwise queries (completed in less than an hour), we used the groupwise interface \changed{for exploration, groups' selection, AL, and groupwise comparison} to get the robot to perform a double backflip from 1,032 preferences, gathered within 34 min.
Fig.~\ref{fig:cases}  presents sample frames capturing the case-study results.

\section{The Simulation Study}
\label{sec:simulationstudy}
\changed{Research evaluating human-feedback settings often utilises models of people's decision-making~\cite{bignold_evaluation_2021,zhang2022time,shi2024interactive,bıyık2022learningpreferencesinteractiveautonomy,arakawa2018dqntamerhumanintheloopreinforcementlearning, shi2025dxhf}.}
In aims of understanding the \emph{general} conditions in which human experts would be able to benefit from our approach, 
we conducted a simulation study modeling a human \changed{\textsc{decision-maker} (DM)} engaged in RLHF.

Running experiments in six environments with the physics engine MuJoCo,
we used five runs, with different seeds, for each setting.
\changed{We modeled \textsc{DM}s with three distinct approaches: standard \emph{pairwise} comparison (\textsc{Pairwise-DM}),\st{and} \emph{groupwise} comparison (\textsc{Groupwise-DM}), and \emph{interactive} groupwise comparison (\textsc{Interactive-DM})}.
\allchanged
\textbf{\textsc{Pairwise-DM}} takes RLHF's traditional approach, wherein one pair of behaviours gets evaluated at a time. It relies completely on the pairs presented to the DM by active learning. 
In \textbf{\textsc{Groupwise-DM}}, an approach permitted only through a data-pre-processing step (see Sec. \ref{sec:cluster}), two groups are compared each time. This method too relies entirely on the samples chosen by AL based on the adapted heuristic with group-variance score.
In \textbf{\textsc{Interactive-DM}}'s groupwise comparison, the comparisons are found not via active learning but interactively, through exploration of the behaviour space, something that is possible only through an exploration interface.
In summary, \textsc{Pairwise-DM} is the baseline approach, \textsc{Interactive-DM} is ours, and \textsc{Groupwise-DM} is is a no-exploration ablation of the latter.

\allsame

We chose six popular robotics tasks for execution in MuJoCo: to
(a) teach \textit{Hopper} to make hops and move, 
(b) teach \textit{Cheetah} to run forward,
(c) teach \textit{Walker} to run forward,
(d) teach \textit{Reacher} to touch a randomly positioned target,
(e) teach the agent in \textit{GridWorld} to move to the goal position, and
(f) teach \textit{MountainCar} to drive to the finish line. \changed{The environments' order of complexity (indicated by the length
of the observation and action vectors) from simplest to most complex is MountainCar (3), GridWorld (9), Reacher (12), Hopper (14), HalfCheetah (23), Walker2d (23).}
The policy was learned via a reward model trained on the feedback from the DM,
and the true reward function acted as the utility function to evaluate how well the trained policies reached these behaviours.

\changed{\subsection{The Decision-Maker.}}
We built DMs by modeling two key steps in \changed{\textsc{Pairwise-DM}, \textsc{Groupwise-DM}, and \textsc{Interactive-DM}}: (1) how users select pairs/groups for comparisons and (2) how they supply their preferences.
The variance score across the ensemble of reward predictors informs the active-learning suggestions.

\begin{itemize}
    \item \changed{\textsc{Pairwise-DM}}: The content with the most variance in predictions between reward predictors gets recommended for comparison. 
    \item \changed{\textsc{Groupwise-DM}}: The group-variance score (see Sec. \ref{par:group_var_score}) dictates the groups for comparison. The suggested groups' maximum size is set to eight behaviours to keep any group from covering too many (an excessive number could make comparisons difficult for humans in real-world settings).
    \item \textsc{Interactive-DM}: Groups to be compared are found by comparing the real average return values of the groups -- something perceptible only to the human who can explore and find comparisons, not through the reward-predictor models. We employed a strategy aimed at an even spread from comparing behaviours with low rewards to comparing ones with high rewards.
\end{itemize}

\newlength{\studywidth}

\setlength{\studywidth}{0.32\textwidth}
\begin{figure*}[!bth]
 \includegraphics[width=.32\linewidth]{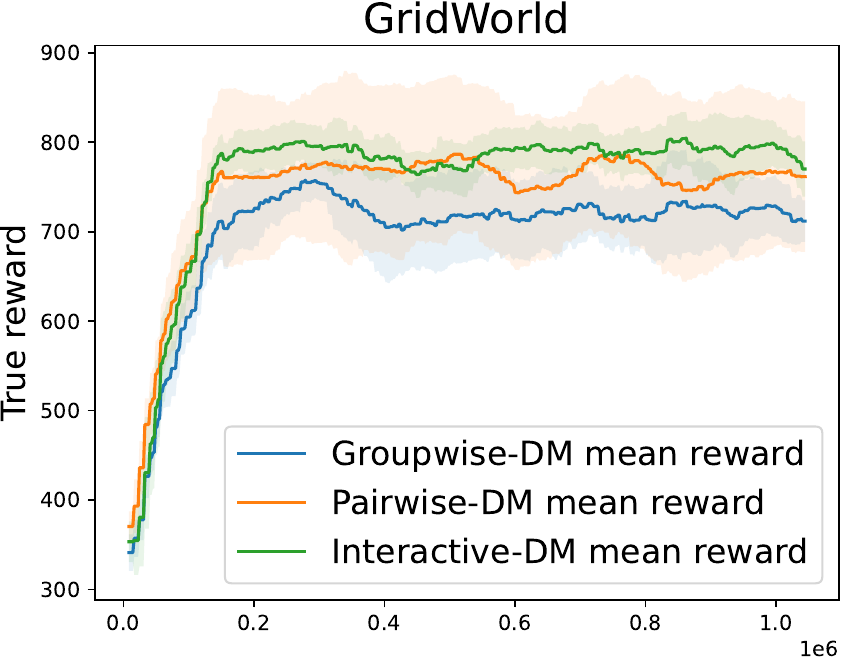} \hfill
 \includegraphics[width=.31\linewidth]{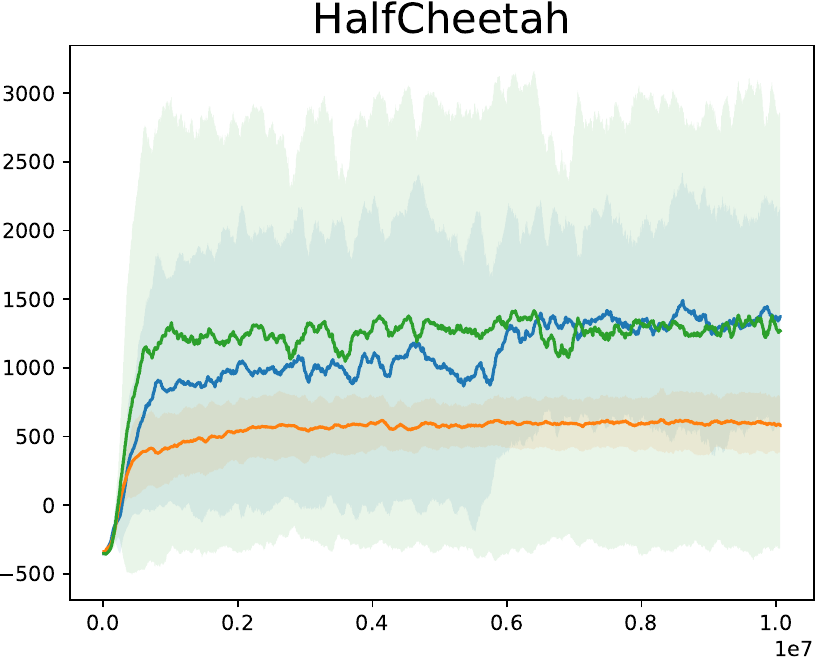} \hfill
 \includegraphics[width=.32\linewidth]{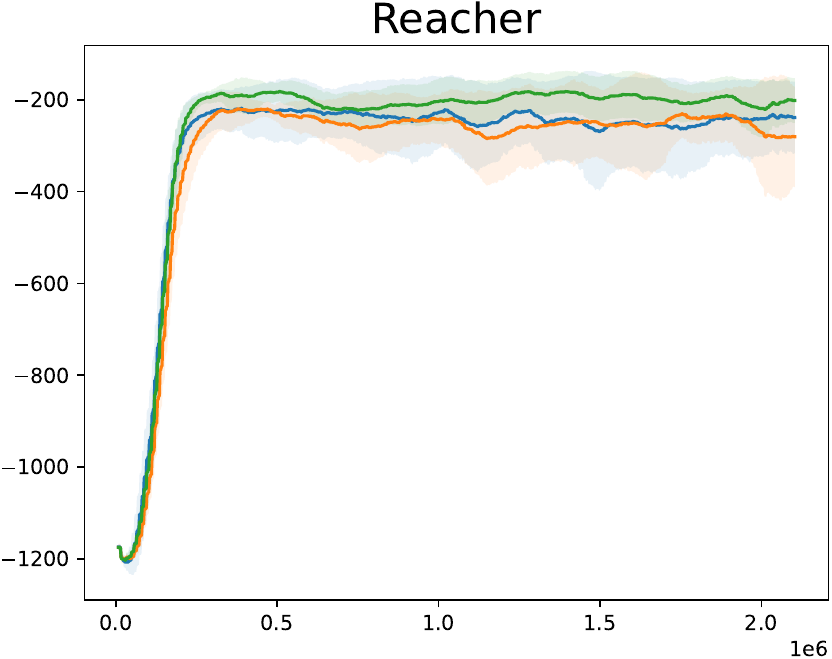}
\vspace{-0.75em}
 \includegraphics[width=.32\linewidth]{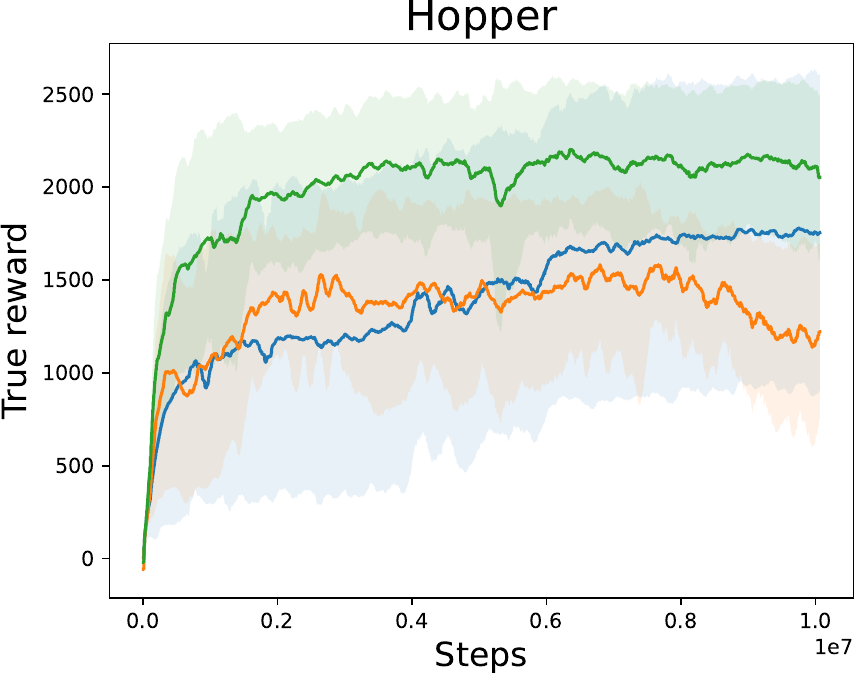} \hfill
 \includegraphics[width=.31\linewidth]{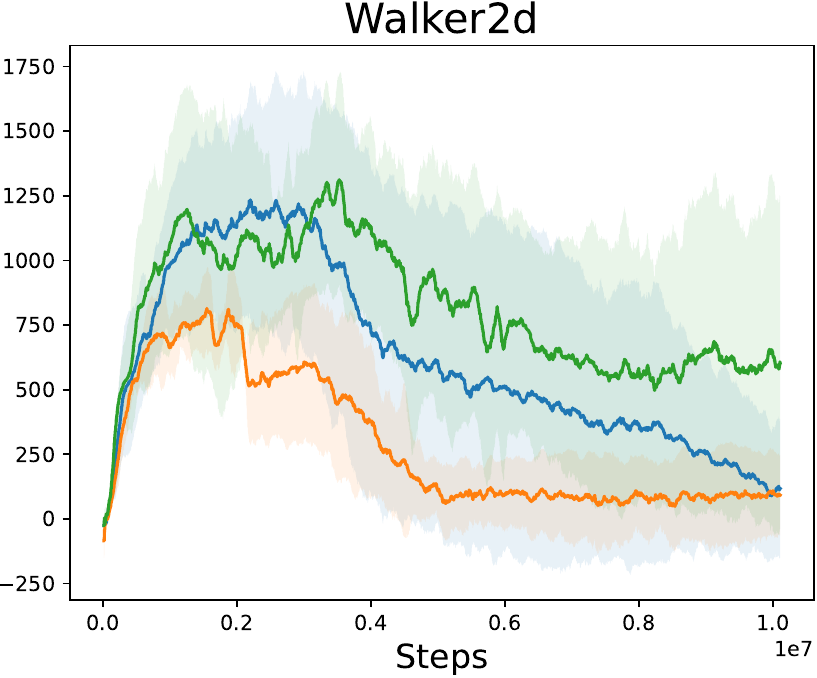} \hfill
 \includegraphics[width=.32\linewidth]{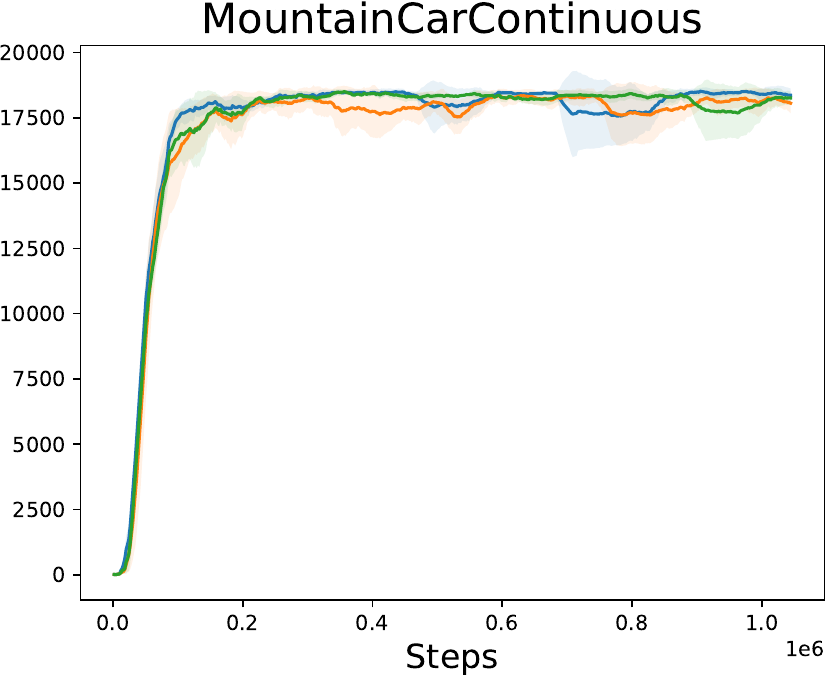}
\vspace{-0.25em}
\caption{\allchanged Simulation-study training logs for six environments, grounded in the true rewards. Each trajectory is the average of five separate runs; the shaded region denotes the confidence interval. In most environments, groupwise active-learning-only comparisons (\textsc{Groupwise-DM}) led to higher true rewards in the final policy than did pairwise comparisons (\textsc{Pairwise-DM}), on average. Interactive groupwise comparisons (\textsc{Interactive-DM}) did the same relative to \textsc{Groupwise-DM}. With all environments, \textsc{Interactive-DM} outperformed \textsc{Pairwise-DM} in terms of the average final reward.\allsame}
\label{fig:synthetic_study_training_logs}
\end{figure*}



We modelled the preferences of the DM based on the noisy user model defined in Eq.~\ref{eq:pref_order}~\cite{Christiano2017DeepRL}.
For \textsc{Pairwise-DM}, preferences were determined from the true rewards plus noise value $\epsilon$. For \textsc{Groupwise-DM} \allchanged and \textsc{Interactive-DM}\allsame, the group whose behaviours' rewards had a higher mean, including $\epsilon$, was considered the preferred one. In the event of significant interference between the rewards of the two groups, comparison was skipped.

\subsection{The Results.}
The rewards achieved by the final trained policies are presented in Table~\ref{tab:reward_comparison}, and Figure \ref{fig:synthetic_study_training_logs} shows the training curves. 
\allchanged
Simpler RL environments (e.g., Gridworld) let us solve properly every time with relatively simple networks (2 fully-connected layers, with 64 neurons) and fairly few feedback instances (400); harder RL problems (e.g., Walker2d) sometimes do not get solved. For this reason, the average training curve of the Walker2d environment slopes downward after 2--3 million steps and HalfCheetah shows very high variance.
\allsame
Generally, \changed{\textsc{Interactive-DM}} outperformed \textsc{Pairwise-DM} across all environments. \allchanged \st{On average, the final reward achieved with \textsc{Groupwise-DM} was 18.3\% higher than that of \textsc{Pairwise-DM} in the scenarios tested.}

Both \textsc{Groupwise-DM} and \textsc{Interactive-DM} yielded a slight decrease in errors made in binary comparisons relative to \textsc{Pairwise-DM}.
Incorrect comparisons grow less likely when the choice is between uniform groups as opposed to single behaviours.
The probability that the perceptions of the rewards are skewed in the wrong direction (such that the wrong result emerges) is lower for two groups with multiple observed rewards than it is for just two behaviours with one reward each. Our user study (reported upon in Sec. \ref{chap:userstudy}) confirmed this finding.

Because the DMs in \textsc{Interactive-DM} often compared more similar groups -- ones unlikely to be suggested by the active-learning technique -- they did more post-selection discarding of groups. Therefore, they made fewer comparisons than in \textsc{Groupwise-DM} with its pure active learning. Therefore, they made fewer comparisons (expressing roughly the same number of preferences as in \textsc{Pairwise-DM}).
The remaining comparisons sufficed for creation of a more robust reward model that had better knowledge of different degrees of `goodness.' The optimal strategy for DM agents and real-world users alike is themselves finding comparisons that cover the range from the most desired to most undesired behaviours well, to guide the agent with a denser reward signal in all the steps needed for reaching a good policy.
\allsame

\begin{table}
    \centering
    \normalsize
    \setlength{\tabcolsep}{3.5pt}
    \renewcommand{\arraystretch}{1.0}
    \begin{tabular}{l|cc|cc|cc}
        \hline
        \multirow{2}*{\makecell[l]{Environment}} & \multicolumn{2}{c}{Pairwise-DM} & \multicolumn{2}{c}{Groupwise-DM} & \multicolumn{2}{c}{Interactive-DM} \\
        ~ & \multicolumn{1}{c}{$M$} & \multicolumn{1}{c}{$+-$} & \multicolumn{1}{c}{$M$} & \multicolumn{1}{c}{$+-$} & \multicolumn{1}{c}{$M$} & \multicolumn{1}{c}{$+-$} \\
        \hline
        Hopper &  1221 & 471 &  1754 & 851 &  \textbf{2053} &  443 \\ 
        HalfCheetah &   579 & 195 &  \textbf{1372} & 799 &  1269 & 1583 \\
        Walker &    92 & 155 &   118 & 268 &   \textbf{603} &  642 \\
        Reacher (-) &  -280 & 110 &  -239 &  78 &  \textbf{-202} &   49 \\
        MountainCar & 18059 & 389 & \textbf{18363} & 263 & 18259 &  261 \\
        GridWorld &   762 &  85 &   712 &  23 &   \textbf{770} &   31 \\
        \hline
    \end{tabular}%
    \caption{\allchanged The average final reward and the standard deviation over five policies for each environment tested
    Active-learning-only groupwise comparison (\textsc{Groupwise-DM}) outperformed standard pairwise comparison (\textsc{Pairwise-DM}) in four of the six environments, and interactive groupwise comparison (\textsc{Interactive-DM}) did the same in every environment. \textsc{Interactive-DM} outperformed \textsc{Groupwise-DM} in four of the environments. 
    \allsame}
    \label{tab:reward_comparison}
    \vspace{-4.0ex}
\end{table}

\allchanged
Delving into the results from training the agents with 400 comparisons (see Table \ref{tab:reward_comparison}) reveals that for four of the six environments \textsc{Interactive-DM} led to higher returns than \textsc{Groupwise-DM} on the true reward. In the other two (HalfCheetah and MountainCarContinuous), \textsc{Groupwise-DM} performed better but not by much; furthermore, the training curve for HalfCheetah 
shows that the average of \textsc{Interactive-DM} reached the maximum reward before a million steps while it took six million steps before the average of \textsc{Groupwise-DM} overtook \textsc{Interactive-DM}. \st{\textsc{Groupwise-DM} gets a larger spread. In every environment, the best training results were through \textsc{Groupwise-DM} but it could fail and produce a comparatively worse final policy.}
\st{The reason is that the simulated users fully follow active learning, without the ability to determine the reasonable selection by themselves.}

In summary, our statistical comparison examining normalised final rewards (scaled such that the interquartile range lies in $[0,1]$) produced three main conclusions:

\textbf{1) Both \textsc{Groupwise-DM} and \textsc{Interactive-DM} outperform \textsc{Pairwise-DM}.}  
The latter reaches a $69.3\%$ higher average reward than \textsc{Pairwise-DM}, with statistically significance ($t=2.684$, $p=9.456e-03$).  
\textsc{Groupwise-DM}'s corresponding advantage over it, $41.3\%$, does not have statistical significance ($t=1.636$, $p=1.072e-01$).

\textbf{2) \textsc{Interactive-DM} and \textsc{Groupwise-DM} yield comparable rewards.}  
Although \textsc{Interactive-DM} yields a $28.0\%$ higher reward, on average, than \textsc{Groupwise-DM}, this difference lacks statistical significance ($t=1.147$, $p=2.560e-01$).

\textbf{3) \textsc{Groupwise-DM} elicits numerous binary preferences while \textsc{Interactive-DM} is more efficient.}  
On average, \textsc{Pairwise-DM} yields $400$ preferences, \textsc{Interactive-DM} $690$, and \textsc{Groupwise-DM} 1,209.  
Groupwise-DM supplies significantly more preferences than \textsc{Pairwise-DM} ($t=2.663$, $p=1.129e-02$), while \textsc{Interactive-DM} exceeds the \textsc{Pairwise-DM} figure somewhat but not significantly ($t=1.033$, $p=3.083e-01$), and it yields significantly fewer preferences than \textsc{Groupwise-DM} ($t=-2.107$, $p=4.175e-02$).

{\subsection{Implications.}
Both \textsc{Groupwise-DM} and \textsc{Interactive-DM} surpass \textsc{Pairwise-DM} in terms of reward. However, \textsc{Interactive-DM} does so more efficiently, from substantially fewer preferences. 

One could object to this interpretation since there are two dependent variables: the number of preferences and the final reward. Therefore, we conducted additional analysis in which we limited the number of preferences that each comparison approach is allowed to return. In that condition -- which is far from realistic, since comparison of groups enables supplying many more preferences -- we again used five training runs per method in all six environments. In this testing, \textsc{Groupwise-DM} reaped worse normalised rewards, with a mean of -0.003 (\textit{STD} 1.238) 
than \textsc{Pairwise-DM}, with a 0.582 mean (\textit{STD} 0.780) (\textit{t} = -2.154, \textit{p} = 3.544e-02). \textsc{Interactive-DM} reached a higher mean, 0.920 (\textit{STD} 0.650), relative to \textsc{Pairwise-DM} (\textit{t} = 1.792, \textit{p} = 7.834e-02) and to \textsc{Groupwise-DM} (\textit{t} = 3.555, \textit{p} = 7.601e-04). Figure \ref{fig:fixed_no_pairs} crystallises the approach-specific distributions of the final returns when the number of preferences is fixed. \textsc{Groupwise-DM} shows the lowest returns on average if we limit the number of preferences to be the same as with \textsc{Pairwise-DM}. That makes sense, in that \textsc{Groupwise-DM}'s key advantage is that it affords returning more preferences. That said, even with the comparison count capped, \textsc{Interactive-DM} proved much better than \textsc{Groupwise-DM} and slightly ($\textit{p} = 7.834e-02$) better than \textsc{Pairwise-DM}.

\begin{figure}
    \includegraphics[width=0.99\linewidth]{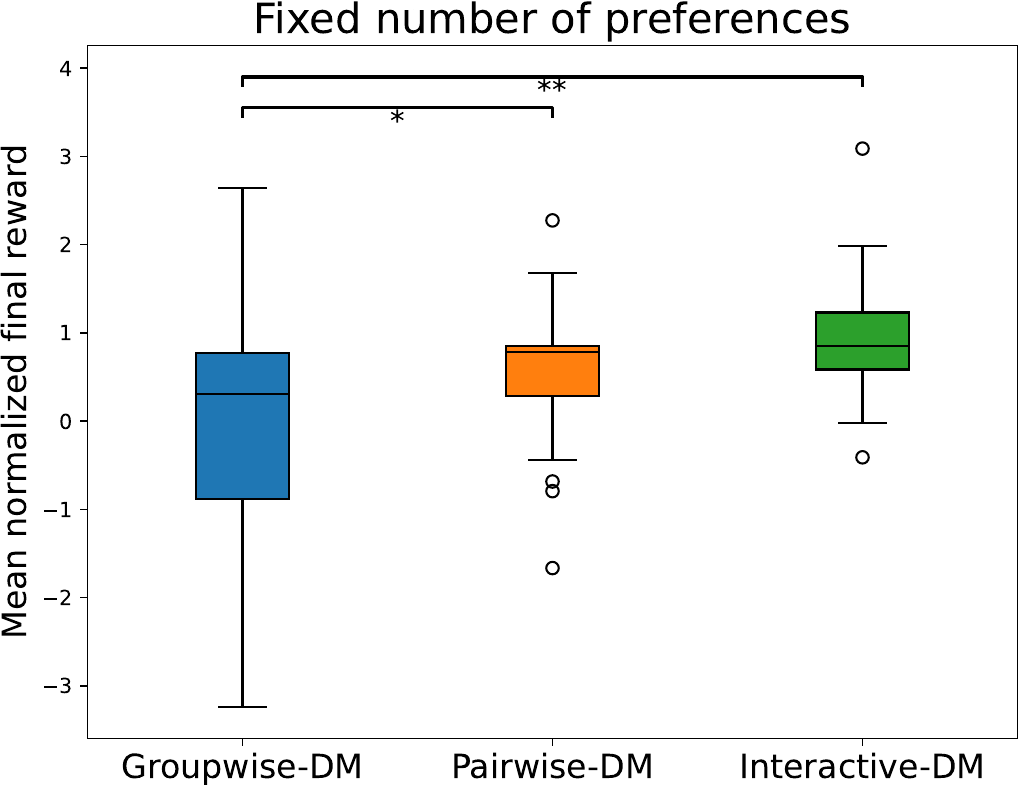}
    \caption{\changed{An unrealistic setting, created for controlled analysis -- when the number of preferences is fixed, \textsc{Groupwise-DM} does not bring higher average rewards than \textsc{Pairwise-DM}. Though groupwise comparison's main benefit is that more preferences can be elicited within the same time, \textsc{Interactive-DM} outperformed both other approaches even with the preference count held constant. The boxes denote the interquartile range, and the black line in each represents the median.}}
    \label{fig:fixed_no_pairs}
\end{figure}

In conclusion, mistakes seem less commonplace when two homogeneous groups rather than two atomic behaviours get mutually compared. The general conditions in which users can benefit from the approach arise when they a) increase the number of pairs judged by using groupwise comparison to their advantage and/or b) find suitable candidate comparisons by exploring the behaviour space, thereby giving more meaningful preferences to the reward model. People who explore might provide fewer comparisons in all but can assure that their feedback gives more information about the target behaviour they want to teach the RL agent.
\allsame

\section{The User Study}
\label{chap:userstudy}

\pgfplotstableread{
Type           PairAvg  PairStdDev    GroupAvg  GroupStdDev
Efficiency     3.3      0.6749485577  4.6       0.5163977795
Usefulness     3.4      0.6992058988  4.6       0.5163977795
{Ease of use}    4.7      0.6749485577  4.2       0.7888106377
} \ExpertRatingTable

\pgfplotstableread{
Name     Pairwise   IDinOldPlotPair     Groupwise    IDinOldPlotGroup
E1       192        3                   417          5               
E2       138        5                   345          2               
E3       225        8                   277          7               
E4       165        2                   261          1               
E5       226        1                   327          8               
E6       198        4                   520          3               
E7       210        6                   348          6               
E8       235        7                   226          4               
E9       202        0                   584          0               
E10      248        0                   502          0               
} \ExpertPrefTable

\newlength{\setupwidth}
\setlength{\setupwidth}{0.22\textwidth}
\newlength{\ratingwidth}
\setlength{\ratingwidth}{0.3\textwidth}
\newlength{\prefwidth}
\setlength{\prefwidth}{0.42\textwidth}

\newlength{\setupheight}
\setlength{\setupheight}{0.739\setupwidth}

%

\begin{figure*}%
\begin{minipage}[t]{\setupwidth}%
\includegraphics[width=\linewidth, trim=5cm 5cm 7.2cm 3cm, clip]{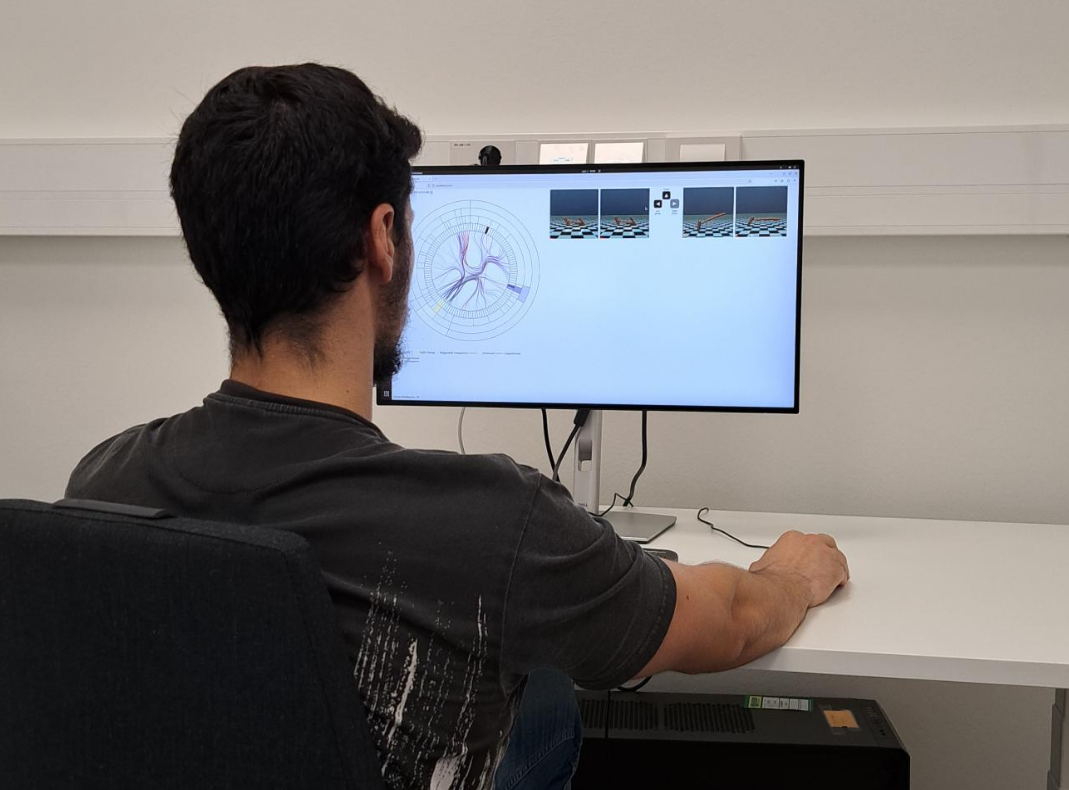}%
\caption{A participant performing the task from a Firefox browser on a desktop computer running Ubuntu, with a 27-inch retina display.}
\label{fig:userstudy}%
\end{minipage}%
\hspace{1.5ex}%
\begin{minipage}[t]{\ratingwidth}%
\resizebox{\linewidth}{!}{%
\begin{tikzpicture}%
\begin{axis}[
width=\ratingwidth,
height=125pt,
ybar,
x=1.5cm,
ymin = 1,
ymax = 5,
enlarge x limits = 0.2,
enlarge y limits = {value=0.2, upper},
bar width = 7pt,
symbolic x coords = {Efficiency, Usefulness, Ease of use},
xtick = data,
xtick pos = lower,
ytick pos = left,
ytick distance = 1,
ylabel = {Ratings},
font = \footnotesize,
legend pos = north west,
legend columns = 2,
]
\addplot+ [error bars/.cd, y dir=both, y explicit, error bar style={black}] table [x=Type, y=PairAvg, y error=PairStdDev] \ExpertRatingTable;
\addplot+ [error bars/.cd, y dir=both, y explicit, error bar style={black}] table [x=Type, y=GroupAvg, y error=GroupStdDev] \ExpertRatingTable;
\end{axis}
\end{tikzpicture}}%
\caption{Participants' ratings on a five-point Likert scale (the vertical lines show the standard deviation). The experts found \textsc{Interactive-UI} more efficient and useful than \textsc{Pairwise-UI}.}
\label{fig:ExpertRating}%
\end{minipage}%
\hspace{1.5ex}%
\begin{minipage}[t]{\prefwidth}%
\resizebox{\linewidth}{!}{%
\begin{tikzpicture}%
\begin{axis}[
width=\prefwidth,
height=157pt,
ybar,
x=0.8cm,
bar width = 7pt,
symbolic x coords = {E1, E2, E3, E4, E5, E6, E7, E8, E9, E10},
xtick = data,
xtick pos = lower,
ytick pos = left,
ymin = 0,
ylabel = {No. of preferences in 21 min},
nodes near coords,
nodes near coords style = {font = \tiny},
font = \footnotesize,
legend pos = north west,
legend columns = 1,
]
\addplot table [x=Name, y=Pairwise] \ExpertPrefTable;
\addplot table [x=Name, y=Groupwise] \ExpertPrefTable;
\legend{\textsc{Pairwise-UI}\ \ \ ,\textsc{Interactive-UI}}
\end{axis}
\end{tikzpicture}}%
\caption{The number of preferences supplied by the users, by comparison type: \textsc{Pairwise-UI} vs. \textsc{Interactive-UI}. The vast majority of the experts (E1--E10) produced more preferences in the time given when using \textsc{Interactive-UI} rather than \textsc{Pairwise-UI}.}%
\label{fig:ExpertNumPreferences}%
\end{minipage}%
\vspace{-5.0ex}
\end{figure*}

We carried out a user study aimed at evaluating the real-world \emph{efficiency}, \emph{usefulness}, and \emph{ease of use} of interactive groupwise comparison (\textsc{Interactive-UI}) versus a baseline of standard pairwise comparison (\textsc{Pairwise-UI}).
The participants were all experts applying RL in their day-to-day work.

\subsection{The Study Design.}
\emph{Participants}:
We recruited 10 expert users (identified as E1--E10) with at least a year's RL experience. The users, three of whom were female, averaged 2.5 years of experience with RL (\emph{SD}=1.58). Eight were already familiar with RLHF, while E4 and E5 encountered it for the first time during our study.

\emph{Experiment procedure}:
We invited the experts to engage in two RLHF sessions, with different tools: one with \textsc{Pairwise-UI} and the other with \textsc{Interactive-UI}.
We chose the hopper test environment, thanks to its easy-to-understand goal for participants.
The experts were instructed that `the centre of the robot is the joint closest to the pointy end. The first priority is for the centre of the robot to move to the right (moving to the left is worse than not moving at all). If the two robots are roughly tied by this metric, then the tiebreaker is how high the centre is.'
\allchanged
With \textsc{Interactive-UI}, the participants were instructed to follow the active-learning suggestions two thirds of the time.

\allsame

\textbf{Experiment design}: 
The study employed a within-subjects design \changed{in which there was one
independent variable, with two levels}: \textsc{Pairwise-UI} and \textsc{Interactive-UI}. We counterbalanced the order of the two conditions.

\textbf{The experiment protocol}:
All sessions were held in a lab setting (shown in Fig.~\ref{fig:userstudy}),
that used a Firefox browser on a Ubuntu desktop platform
with a 27-inch retina
display $(2560\times1440, 60~fps)$
and a commodity GPU (NVIDIA GeForce RTX~2080).
Each expert
began with 30 min of training to guarantee an opportunity for familiarity with the usage of the tools. The first 10 min consisted of a tutorial giving an introduction to the tools.
Then, participants freely tested the environments and played with the tools. 
After the training came the two RLHF sessions, each lasting about 35 min. The time for giving feedback in RLHF was fixed, with all participants completing seven three-minute rounds of feedback (for 21 min of user work per tool). After each round, the models were retrained for about two min, and new videos were generated for user analysis. Each tool recorded the number of preference choices returned and logged training performance as measured by the true reward.
After completing both sessions, which a researcher observed, the experts completed a questionnaire and the researcher conducted a post-experiment interview lasting about 20 min. 
Each study lasted roughly 2 hours.
Participants received a 30-euro gift voucher as compensation for their time.

\subsection{Results.}
\emph{Overall ratings}:
Per the overall ratings from the questionnaire, outlined in Figure~\ref{fig:ExpertRating},
\textsc{Interactive-UI} comparison fared \allchanged\st{significantly}\allsame better than \textsc{Pairwise-UI}'s
for \emph{efficiency} \allchanged\st{(paired t-test: $p < 0.01$)}\allsame
and \emph{usefulness}\allchanged\st{(paired t-test: $p < 0.05$)}\allsame.
The participants rated
\textsc{Pairwise-UI}
higher for \emph{ease of use}\allchanged\st{,
however, that difference is not statistically significant}\allsame.


\emph{Efficiency and the feedback's accuracy}:
For each approach, we tabulated the number of preferences that the participants supplied (see Fig.~\ref{fig:ExpertNumPreferences}). All experts \allchanged but one \allsame offered more preferences with \textsc{Interactive-UI} than \textsc{Pairwise-UI}. On average, users supplied 86.7\%  more in the same span of time when using \textsc{Interactive-UI}. 
We also factored in the number of mistakes that the experts made in their preference indications, based on the true reward known for the task. Their error rate was generally lower with \textsc{Interactive-UI} (10.8\% of preferences) as compared to \textsc{Pairwise-UI} (12.8\%).
In summary, \textsc{Interactive-UI} yields more preferences in total, along with an error rate lower than \textsc{Pairwise-UI}'s.
\changed{
Adhering to the initial instructions fairly well, participants obtained comparison candidates from the exploration view about a third of the time. They often succeeded in producing lower error rates, more preferences, and better policies.}

\emph{Quality of the trained policies}:
The average reward from \textsc{Interactive-UI} was 1043,
which is 60.9\% higher overall than the reward value from \textsc{Pairwise-UI}, 648. \allchanged\st{However, there is no statistically significant difference between the two approaches (non-parametric sign test: $p = 0.172$).
%
This lack of significance may be due to the reliance on the quality of human feedback, which can be somewhat random.} However, the sample in the user study was quite small, not permitting claims of statistical significance. \allsame
In 7 of the 10 sessions, \textsc{Interactive-UI} produced better policies than \textsc{Pairwise-UI} after the set feedback-provision time. From those seven sessions, the hopper
learned to move forward, bringing higher scores than the \textsc{Pairwise-UI} condition. \textsc{Interactive-UI} yielded the four best policies, with the top one having a true reward of over 2500.  
\changed{\st{In one session, however, neither approach performed well. The expert participating in that session (E2) had some difficulty in ascertaining preferences based on the short clips, in both settings.
In 2 sessions (E6 and E10), \textsc{Pairwise-UI} was significantly better than \textsc{Interactive-UI}. In the interview, E6 stated, `The tool [\textsc{Interactive-UI}] was neat; however, I was probably a bit tired when making comparisons and potentially made some mistakes in judging the trajectories'; likewise, E10 stated that early on with that tool `I might have misclicked a lot, giving incorrect preferences, since I used the keyboard with my left hand and I'm not left-handed.'}} 

\allchanged
\allsame

\emph{Experts' remarks}:
\allchanged
\st{In the interviews, participants indicated
that \textsc{Interactive-UI} allows for quicker feedback by clustering behaviours. Some participants had also found it effective for control and selection. The tool's ability to target specific groups on the basis of patterns was appreciated, though a few mentioned that it occasionally led to mistakes when grouping similar yet distinct behaviours together. Despite this shortcoming, the users generally found it efficient for selecting the behaviours they preferred.
Regarding the tool's utility, most participants stated that \textsc{Interactive-UI} helped them obtain clearer understanding of the behaviour space across organised groups of videos. 
They appreciated being able to trace the video clips in the exploration view and see their earlier feedback. Experts found this a more exploratory experience
than usual, one
that aids in tracking one's progress and seeing which comparisons have not yet been made. Additionally, one participant mentioned that the tool's recommendations rendered it easier to select better groups for comparison.}
All but one of the participants found \textsc{Interactive-UI} efficient to work with (for example, `It is efficient because I have a global overview,' `Yes, more control over similar cases,' and `Yes, I can control what I wanna compare'). Only E1 deemed \textsc{Pairwise-UI} efficient, thanks to its simplicity (`It's efficient because it only provides three options'), while declaring \textsc{Interactive-UI} inefficient since `although I get more detailed visualisation and more options, it is not efficient [when] compared to only three options.'
Seven users reacted negatively to the \emph{information} from \textsc{Pairwise-UI}, and the others remained neutral; for example, E3 stated,
`Local, limited, does the job but minimally, so it's slightly boring and repetitive.'
All of the experts gave favourable assessments of the information supplied by \textsc{Interactive-UI}: `More inclusive of the bigger picture, more complete visualisation of the data coverage so far,' `It was useful to be able to see previous preferences and where in the `trajectory space' the clips came from,' `The tool['s] suggestions [were] nice, to select better videos,' and so forth

Although three experts expressed neutral opinions about \textsc{Pairwise-UI}'s controllability and the other seven responded negatively (`It is simple but not helpful in exploring the behaviours,' `It does not [let one] see a lot of behaviours,' `The pairwise tool didn't allow much exploration; I imagine that mistakes in the pairwise tool would be quite costly,' and so forth), all but one stated that they sensed more control with \textsc{Interactive-UI}. 


That said, \textsc{Interactive-UI} was experienced as harder to use. For example, one participant stressed that the way of comparing clips in this tool `needs several rounds to get used to.'
Even E7, who racked up the highest scores across all experiments, said of \textsc{Interactive-UI} that `I could do more comparisons at the same time and choose better videos but it was more cognitively demanding' while \textsc{Pairwise-UI} made it `quick to select the best video and less cognitively demanding.' Indeed, his score in the pairwise experiments was the second highest.

\allsame


\emph{The experts' input in summary}:
We conclude that \textsc{Interactive-UI} offers efficient, flexible functionality but requires more cognitive effort. On the other hand, \textsc{Pairwise-UI} is simpler, with limited exploration and control capabilities. These tradeoffs emphasise the importance of considering task complexity and user cognitive load in future iterations of comparison tools.

\allchanged
\section{Discussion}
Notwithstanding the clear advantages ushered in with the exploration view and tools for groupwise comparison, we recognise some limitations.


Granting users the power to provide quality feedback in larger quantities brings a danger of drawing in more meaningless feedback and added noise if they pay little attention or work in very broad strokes only. The user must strike a careful balance so as to be quick yet still provide variety-rich, accurate feedback. Our study with real human users spotlighted this factor. Although giving them a set time for both interfaces, to afford fair comparison between the two, might be a somewhat unrealistic constraint, properly judging the relative advantages of the interface demanded it. These conditions revealed that some experts
could mesh \textsc{Interactive-UI} with their way of working and train much better policies while others gave less valuable feedback -- so much so that their policies' rewards were lower than with the standard interface even though they stated more preferences in absolute terms. Skills and effort are required.
Cognitive load, stress, and so forth rear their head here. In the interviews, some users cited the mental burden in the interactive visualisation as greater than with a tool that does not allow for exploration. The questionnaire data echo this sense of higher perceived cognitive load: \textsc{Interactive-UI} received lower average scores for only ease of use, relative to the pairwise baseline.

The design's scalability imposes further limits. Although our system enables user agency and exploration, its radial design faces restrictions and might not scale well to \changed{\st{extremely large (e.g. LLM)}}behaviour spaces \changed{where two behaviours can differ on many thousands of dimensions (e.g., LLaMa~3 applies a 4,096-dimension token-embedding vector for each token)}.
Although the specific design we deployed serves smaller cases better, enabling its methods for user agency and exploration could still benefit handling other cases too.

This design is limited also in that the visualisation of the behaviour space remains quite abstract until a behaviour or group of them gets selected. Therefore, a feature pointing out representative behaviours within a group could be worthwhile: this should let users see at a glance what characterises the group before they choose it. Another idea for future work came from one of the expert users, who suggested showing a preview of the clips next to the mouse upon hovering over a behaviour in the exploration view. Although the current design does not support these features, they could be added, advancing the tool greatly in future design iterations.

In theory, comparing groups rather than individual behaviours could increase error rates, since outliers within a group may distort the labels generated. Empirically, we observed the opposite. Further research is needed to identify the conditions under which this error-reduction tendency occurs and whether it reverses in some other circumstances.


There are various avenues for developing design features that enhance the approach further. For now, the most obvious safeguards for success in applying it are to be sure the sample from the behaviour space is not overly large (sampling fewer than 200 behaviours per round is best) and work with well-trained, attentive users.
\allsame

\section{Conclusions}

Our central contribution is a novel approach for interactive groupwise comparison
of behaviours
for RLHF.
\changed{Evaluations of its interactive visualisation, which includes a hierarchical radial chart and edge bundling to aid in exploring and analysing behaviours for comparison, attest to the trained policy returns improving by 69.34\% with respect to the baseline.}
Its efficiency benefits were highly evident from the evaluation study carried out with experts, in which the number of preferences elicited within a given time frame rose by 86.7\%. \changed{Moreover, the interactive approach reduced error rates (incorrect preference input).}
Expert interviews support concluding that the new approach empowers users to control their exploration and gain comprehensive understanding of the context.

There are inherent tradeoffs to consider when reflecting on interactive groupwise comparison against the backdrop of standard pairwise comparison.
\changed{\st{Firstly, o}O}ur approach offers high efficiency
but may require longer training, to familiarise users
with its visual-analysis features.
Also, it demands more cognitive investment during use.
However, users exposed to it
emphasised its efficient functioning and flexibility.
\allchanged
\st{Second, the standard pairwise comparison approach may be more flexible when dealing with various types of data. For example, it can handle the comparison between 2 text sentences to tune preferences for a large language model. 
Groupwise comparisons for text
is an interesting avenue for future research.}
Through such advances, visualisation research has much to offer processes for training AI models (RLHF and others) by designing interfaces with room for
strong user agency and that help users make full use of their cognitive capabilities.\allsame

\section{Additional Information}

\textbf{Correspondence:} Prof. Antti Oulasvirta, \emph{antti.oulasvirta@aalto.fi}, is the corresponding author.

\textbf{Data Availability:}  
The codebase for reproducing the experiments is available at \url{https://github.com/jankomp/interactive_rlhf}.  
Data from the user study can be furnished by the corresponding author upon reasonable request.

\textbf{Conflicts of Interest:}  
The authors declare that they have no conflict of interest.

\textbf{Funders:}
JK, SD, and AO received support from the Research Council of Finland (FCAI: 328400, 345604, 341763; Subjective Functions: 357578) and the ERC (AdG project Artificial User: 101141916). TW was supported by the Swedish e-Science Research Centre \emph{SeRC}. 

\textbf{Ethics:}
Local regulations do not require formal ethics review.


\bibliographystyle{eg-alpha-doi}  
\bibliography{reference}     


\newpage

\end{document}